# Mining the Web for Pharmacovigilance: the Case Study of Duloxetine and Venlafaxine

Abbas Chokor, MS[1], Abeed Sarker, PhD[1], Graciela Gonzalez, PhD[1]
[1]Arizona State University, Tempe, AZ

**Abstract**

*Adverse reactions caused by drugs following their release into the market are among the leading causes of death in many countries. The rapid growth of electronically available health related information, and the ability to process large volumes of them automatically, using natural language processing (NLP) and machine learning algorithms, have opened new opportunities for pharmacovigilance. Survey found that more than 70% of US Internet users consult the Internet when they require medical information. In recent years, research in this area has addressed for Adverse Drug Reaction (ADR) pharmacovigilance using social media, mainly Twitter and medical forums and websites. This paper will show the information which can be collected from a variety of Internet data sources and search engines, mainly Google Trends and Google Correlate. While considering the case study of two popular Major depressive Disorder (MDD) drugs, Duloxetine and Venlafaxine, we will provide a comparative analysis for their reactions using publicly-available alternative data sources.*

**Introduction**

Despite the continuous surge in developing drug therapies for major diseases, only 30% to 50% of patients has shown improvements, while approximately 7% of hospitalized patients experience adverse drug reactions[1] (ADR). Adverse reactions caused by drugs following their release into the market are among the leading causes of death in many countries. Pharmacovigilance systems, also known as ''early warning'' systems have monitored ADRs since 1960s by health agencies and societies including the World Health Organization[2] (WHO). In the United States, The Food and Drug Administration (FDA) is the federal agency that is responsible for the protection and promotion of public health including the regulation and supervision of drugs[3]. Before its release to the market, a pharmaceutical product must go through rigorous testing, in clinical trials, for short-term safety and efficacy on a very limited number of subjected which were carefully chosen using restrictive inclusion an exclusion criteria[4]. However, the various limitations of clinical trials prevent a comprehensive assessment for the consequences of using a particular drug before its release[5].

In order to maximize the benefic effects, while reducing the risk of adverse reactions due to toxicity or sub therapeutic dosing, scholars and practitioners have expanded their pharmacovigilance efforts using different approaches. For instance, the rapid growth of electronically available health related information, and the ability to process large volumes of them automatically, using natural language processing (NLP) and machine learning algorithms, have opened new opportunities for pharmacovigilance that could address some of the above- mentioned limitations[6]. A recent survey found that more than 70% of US Internet users consult the Internet when they require medical information[7]. The data created by people who thus seek information offers an unprecedented opportunity for applications to monitor and improve the quality of life of people with a variety of medical conditions[8]. Although pharmacovigilance on Twitter data has been tackled by several studies[6, 9-13], such data can be used in a variety of ways to the betterment of medical research and practice.

Major depressive Disorder (MDD) is one of the most common mental disorders worldwide, affecting negatively the daily living for extended periods of time[14]. MDD is ubiquitous, resulting in almost 16 million disabilities in the United States[15] and 350 million people worldwide[16] each year. While major depressive disorder can develop at any age, it affects approximately 14.8 million American adults, or about 6.7 percent of the U.S. population age 18 and older, in a given year[17]. Moreover, studies have shown that MDD is more prevalent in women than in men[18]. In addition, people with depression are four times as likely to develop a heart attack as those without a history of the illness. After a heart attack, they are at a significantly increased risk of death or second heart attack[19]. Alongside the evolution and the increase in MDDs, scientists have worked avidly on developing antidepressant drugs. Historically, the treatment of MDD has mainly entailed several classes of medications, including tricyclic antidepressants, benzodiazepines, and selective serotonergic reuptake inhibitors[20] (SSRIs). However, the widespread use of SSRIs vis-à-vis older antidepressant classes is predominantly because of their greater ease of use and more favorable side effect profile[21]. More recently, serotonergic noradrenergic reuptake inhibitors (SNRIs), another group of evidence-based pharmacological interventions for the MDD treatment, has been shown to be more efficacious than SRRIs[22-26]. Within SNRIs, previous studies have reported the superiority of duloxetine and venlafaxine for MDD treatment[27-28].

Both duloxetine and venlafaxine, sold under the brand names Cymbalta and Effexor respectively among others, are SNRIs designated for the treatment of MDD. Although they have similar mechanisms of actions, duloxetine has a more balanced affinity for both serotonin and norepinephrine transporters, whereas venlafaxine has a higher affinity for serotonin than norepinephrine transporters[29-30]. From one side, duloxetine inhibits the reuptake of serotonin and norepinephrine (NE) in the central nervous system. Duloxetine increases dopamine (DA) specifically in the prefrontal cortex, where there are few DA reuptake pumps, via the inhibition of NE reuptake pumps (NET), which is believed to mediate reuptake of DA and NE[31]. However, duloxetine has no significant affinity for dopaminergic, cholinergic, histaminergic, opioid, glutamate, and GABA reuptake transporters and can therefore be considered to be a selective reuptake inhibitor at the 5-HT and NE transporters[32-33]. From the other side, venlafaxine blocks the transporter "reuptake" proteins for key neurotransmitters affecting mood, thereby leaving more active neurotransmitters in the synapse, such as: serotonin, norepinephrine, and dopamine[34].
In this paper, we presented our effort in analyzing and comparing the trends through the web content mining for duloxetine and venlafaxine. In addition to collecting and analyzing the data from *Twitter*, we also investigated the data from different sources, such as *Google Trends* and *Google Correlate*. We anticipate that our work could help highlight different data resources that can be applied in medical research for different drugs and classes

**Methods**

*Google Trends:* Google Trends is a public web facility of Google Inc. that provides a time series index of the volume of queries users enter into Google. The query index is based on query share: the total query volume for the search term in question within a particular geographic region divided by the total number of queries in that region during the time period being examined. The maximum query share in the time period specified is normalized to be 100 and the query share at the initial date being examined is normalized to be zero[35]. The horizontal axis of the main graph represents time (starting from 2004), and the vertical is how often a term is searched for relative to the total number of searches. This query index data is available at country, state, and metro level for the United States and several other countries. On August 5, 2008, Google launched Google Insights for Search, a more sophisticated and advanced service displaying search trends data. On September 27, 2012, Google merged Google Insights for Search into Google Trends[36]. Google classifies search queries into about 30 categories at the top level and about 250 categories at the second level using a natural language classification engine. After searching for a term, Google Trends provides a list of the most frequently searched with the term you entered in the same search session, within the chosen category, country, or region.

To identify the reactions of patients on duloxetine and venlafaxine, we compared Trends of duloxetine (Cymbalta) and venlafaxine (Effexor) in the United States. Then, we analyzed the graph by inspecting the term's popularity over time. We also examine different points on the graph where the letters represent news articles that might indicate why a certain term is spiking during that time period. Next, we found the related top searches for both drugs.

*Google Correlate:* Google Correlate is an experimental new tool on Google Labs which enables you to find queries with a similar pattern to a target data series. The target can either be a real-world trend that you provide (e.g., a data set of event counts over time) or a query that you enter[37]. Google Correlate contains web search activity data from January 2003 to present and updated weekly. It computes the Pearson Correlation Coefficient (r) between your target query and the frequency time series for every query in Google database[38]. Correlation coefficients range from r=-1.0 to r=+1.0.

To inspect the related interests of patients on duloxetine and venlafaxine, we searched for the correlated terms with duloxetine (Cymbalta) and venlafaxine (Effexor) in the United States. Then, we extracted and classified the related terms with the highest correlation coefficient (i.e. closest to r=1.0).

*Twitter Mining:* To meet the objectives of this study, Twitter data was used to inspect the sharing and posting health related information for duloxetine and venlafaxine. The Diego Lab provides a manually annotated corpus for drugs related comments from mining the Twitter microblogging platform[39]. After filtering the target drugs from the corpus, we found 5640 duloxetine (Cymbalta) tweets and 6392 venlafaxine (Effexor) tweets between April 2013 and September 2015. First, we preprocessed the comment texts by lowercasing the characters and stemming all the terms using the Porter stemmer[40]. Then, we removed the stop words and extract the tokens using the Regexp tokenizer[41].

Next, we identified the words that are most informative about the topic and genre of the tweets by calculating the frequency distribution of the tokens. Moreover, we extracted the main collocations from the tweets for both drugs using the Collocations finder[42]. Later, we conducted a sentimental analysis for the preprocessed tweets to inspect the satisfaction of patients with the studied drugs. For this analysis, we attempted to incorporate a score that represents the general sentiment of a comment (as the normalized sum of all the terms in the comment). Each word in a comment is assigned a score and the overall score assigned to the comment is equal to the sum of all the individual term sentiment scores, normalized by the length of the sentence in words. To obtain a score for each term, we used the Stanford sentimental analysis lexicon[42]. The overall score a sentence receives is therefore a floating point number with the range of -1 to 1.

**Results and Discussion**

*Google Trends Analysis*: After searching for Cymbalta and Effexor on Google Trends, the results showed the popularity variation for the inspected drugs. Figure 1 below shows that patients tend to be more interested into Cymbalta than Effexor. Furthermore, a consideration of the main events presented by letters reveals the reaction of patients with the drug milestones. For example, the approval of Cymbalta for muscle pain and chronic joint had a good impact on the web users for two years before dropping down again starting 2013.

**Figure 1.** Google Trends for Cymbalta and Effexor.

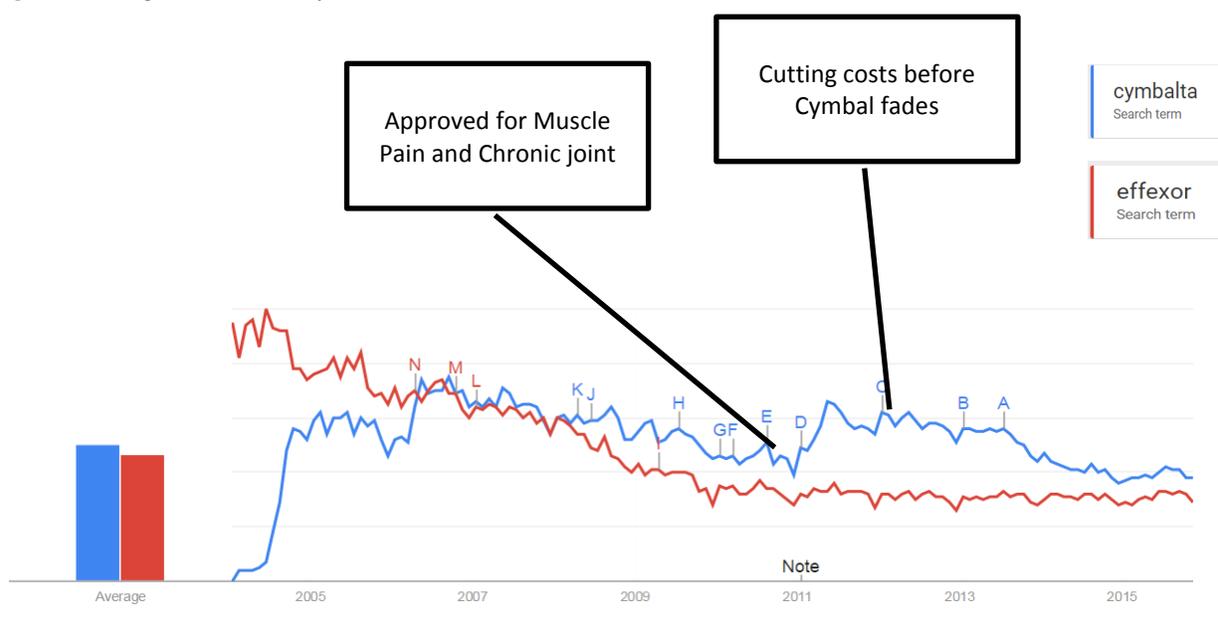

In addition, table 1 lists the top five related queries for each one of the investigated drugs. The related queries disclose the high interest of web users with the side effects of both drugs mainly. Moreover, users were interested into the withdrawal of drugs and their effects. Out of the top queries of Cymbalta is "pain" and that is what can be explained by the fact that FDA Approves Cymbalta was also approved in 2011 for Chronic Musculoskeletal Pain. Wellbutrin, a norepinephrine-dopamine reuptake inhibitor (NDRI) antidepressant drug, also appeared in the top queries related to Effexor.

**Table 1.** Google Trends Top Related Queries.

| Cymbalta Top Related Queries | | Effexor Top Related Queries | |
|---|---|---|---|
| Side effects Cymbalta | 100 | Effexor XR | 100 |
| Cymbalta generic | 20 | Side effects Effexor | 95 |
| Cymbalta withdrawal | 20 | Effexor withdrawal | 30 |
| Effects of Cymbalta | 20 | Wellbutrin | 20 |
| Cymbalta pain | 15 | Effexor generic | 20 |

The relationship between Effexor and Wellbutrin (also known as Bupropion) can be explained by the questions and posts of users on the web asking on the effects of taking both drugs together. For example, the following conversation is extracted from Drugs.com Questions and Answers:

In 22 Oct 2014, *Jlamar* asked "Can I take Venlafaxine and Wellbutrin together? I have taken these drugs at different in the past?" The next day, *Evrclr29* answered "Wellbutrin may make you more anxious so you might need to up your Effexor to make up for it".

*Google Correlate Analysis*: After searching for Duloxetine and Venlafaxine in Google Correlate, the top queries with a similar pattern to drugs data series were identified. Figures 2a and 2b illustrate the correlations geographically in the US with the top queries.

**Figure 2.** Google Correlate Maps for (a) Duloxetine and (b) Venlafaxine.

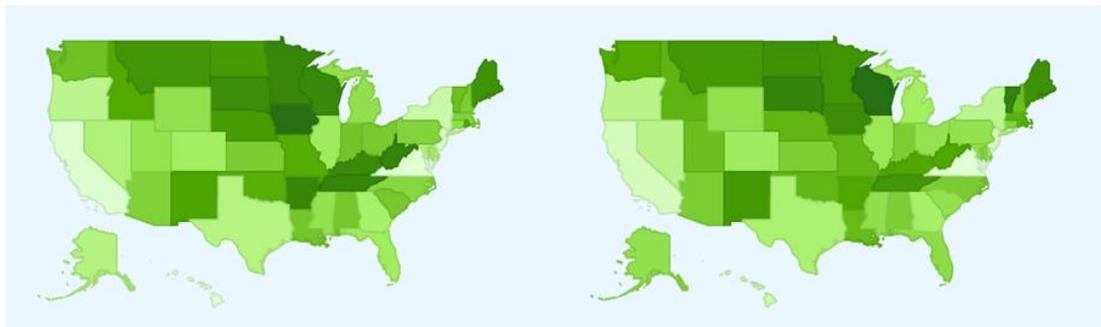

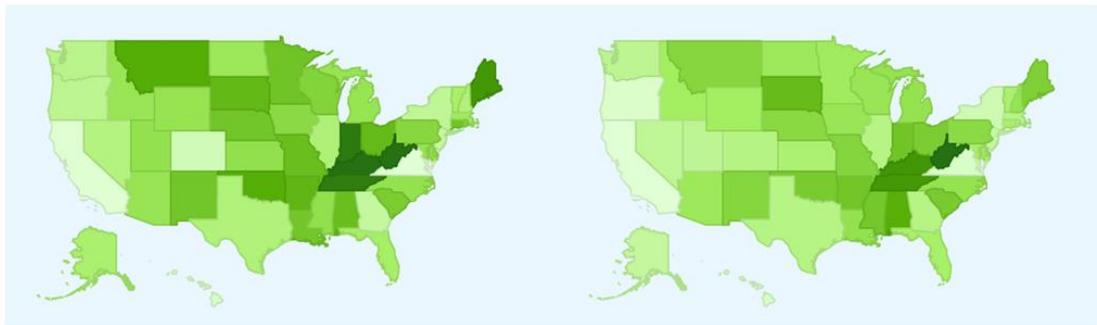

Table 2 and 3 below classify the top 20 queries for duloxetine and venlafaxine respectively into syndrome/disease/potential ADR, potential drug interactions, uses, and others. The results show that Google correlate can extract the top related queries and highlight some possible ADR that have not been announced or declared based on the users' web searches.

*Tweets Analysis:* An illustration of the main tokens is shown in figures 3(a) and 3(b) from the related tweets mentioning duloxetine and venlafaxine respectively. Figure 4 compares the results of sentimental analysis for both drugs. Although both drugs have close sentimental scores, Cymbalta users tend to be more satisfied with the drug as a result of specific events. The reasons behind these results alongside the events and milestones of both drugs will be investigated in future studies. Moreover, the top collocations are presented in table 4.

**Table 2.** Google Correlate Top Queries for Duloxetine

| Duloxetine Queries | Correlation Factor | Syndrome/Disease/ Potential ADR | Potential Drug Interactions | Uses | Others |
|---|---|---|---|---|---|
| pancreatitis | 0.8967 | X | | | |
| lantus | 0.8947 | | X | | |
| ziprasidone | 0.8919 | | X | | |
| generic names | 0.8821 | | | | X |
| ropinirole | 0.8793 | | X | | |
| tachycardia | 0.8766 | X | | | |
| benzodiazepines | 0.8714 | | X | | |
| bipap | 0.8696 | X | | | |
| quadrant pain | 0.8696 | | | X | |
| requip | 0.8693 | | X | | |
| chronic pancreatitis | 0.8634 | X | | | |
| dosing | 0.8634 | | | | X |
| upper quadrant pain | 0.8621 | | | X | |
| malformation | 0.862 | X | | | |
| lower quadrant pain | 0.8615 | | | X | |
| epididymitis | 0.8606 | X | | | |
| right upper quadrant pain | 0.8598 | | | X | |
| sinus tachycardia | 0.8595 | X | | | |
| pneumonia vaccine | 0.8593 | | X | | |
| incisional hernia | 0.8566 | X | | | |

**Table 3.** Google Correlate Top Queries for Venlafaxine

| Venlafaxine Queries | Correlation Factor | Syndrome/Disease/ Potential ADR | Potential Drug Interactions | Uses | Others |
|---|---|---|---|---|---|
| bupropion | 0.9181 | | X | | |
| citalopram | 0.9178 | | X | | |
| sumatriptan | 0.9081 | | X | | |
| paroxetine | 0.9028 | | X | | |
| serotonin syndrome | 0.8957 | X | | | |
| granuloma annulare | 0.8913 | X | | | |
| steatohepatitis | 0.8898 | X | | | |
| annulare | 0.8877 | X | | | |
| sertraline | 0.8793 | | X | | |
| fluoxetine | 0.8785 | | X | | |
| amitriptyline | 0.8759 | | X | | |
| carbs in apple | 0.871 | | | | X |
| i love a rainy night | 0.8706 | | | | X |
| ostomy | 0.8694 | X | | | |
| diltiazem | 0.8673 | | X | | |
| ove a rainy night | 0.8638l | | | | X |
| propranolol | 0.8627 | | X | | |
| lantus insulin | 0.8626 | | X | | |
| low back pain | 0.8603 | | | X | |
| mccarthy movies | 0.8586 | | | | X |

**Figure 3.** Word Cloud Presentations for (a) Duloxetine and (b) Venlafaxine.

**Figure 4.** Sentimental Analysis Results for Effexor and Cymbalta

**Table 4.** Duloxetine and Venlafaxine Collocations

| Duloxetine | | Venlafaxine | |
|---|---|---|---|
| Cymbalta | help | Take | effexor |
| Hurts | cymbalta | Side | effects |
| Depression | hurts | Effexor | withdrawal |
| Need | cymbalta | Taking | effexor |
| Take | cymbalta | Effexor | effexor |
| Cymbalta | commercial | venlafaxine | withdrawal |
| Help | depression | Feel | like |
| Help | cymbalta | Took | effexor |
| Help | http | Brain | zaps |
| Cymbalta | depression | Effexor | made |
| Side | effects | Effexor | withdrawals |
| Cymbalta | can | Dose | effexor |
| Cymbalta | cymbalta | Cold | turkey |
| Can | help | Last | night |
| Cymbalta | helps | Take | venlafaxine |
| Like | cymbalta | withdrawal | symptoms |
| Taking | cymbalta | Coming | effexor |
| Hurt | cymbalta | Get | effexor |
| Get | cymbalta | Effexor | worst |
| Cymbalta | withdrawal | Effexor | like |
| Feel | like | Like | effexor |
| Cymbalta | doesn | venlafaxine | effexor |
| Cymbalta | right | Effexor | today |
| Depression | hurt | Side | effect |
| Think | cymbalta | Think | effexor |
| Depression | cymbalta | Effexor | makes |
| Lyrica | cymbalta | Put | effexor |
| Doctor | cymbalta | Effexor | works |
| Cymbalta | commercials | Effexor | really |
| Cymbalta | need | Effexor | dose |

**Conclusion**

We have shown how web data can facilitate medical research, providing an overview of the state-of-the art in this area of Pharmacovigilance. While previous studies have relied mainly on Twitter and Medical forums and website to monitor the drugs and find ADR, we exposed the importance of information that we can extract from alternative data sources such as Google Trends and Google Correlate, while investigating the case study of duloxetine (Cymbalta) and venlafaxine (Effexor). This paper offers the researchers alternative tools to address medical questions using web data. We were, however, only able to analyze the trends and data from different sources separately. The main reason behind that is the large data imbalance associated with data collected from between Twitter and Google. In the future, we will also analyze the time series of events affecting the trends and correlation. Such analysis will aim to predict and forecast the trends of the drugs. We will also study the variation in drugs sales and prices and try to investigate the interaction between the satisfaction of patients and drug prices from one side and the sales from the other side.

# References

1. S. D. Undevia, G. Gomez-Abuin, and M. J. Ratain, "Pharmacokinetic variability of anticancer agents," *Nature Reviews Cancer*, vol. 5, pp. 447-58, 2005.
2. Bruno HS, Bruce MP. Detection, verification, and quantification of adverse drug reactions. BMJ 2004;329:44–7.
3. U S Food and Drug Administration Home Page.
4. Jiang K, Zheng Y. Mining twitter data for potential drug effects. In Advanced Data Mining and Applications 2013 Jan 1 (pp. 434-443). Springer Berlin Heidelberg.
5. Harpaz R, DuMouchel W, Shah NH, Madigan D, Ryan P, Friedman C. Novel data- mining methodologies for adverse drug event discovery and analysis. Clin Pharmacol Ther 2012; 91(3):1010–21.
6. Sarker A, Ginn R, Nikfarjam A, O'Connor K, Smith K, Jayaraman S, Upadhaya T, Gonzalez G. Utilizing social media data for pharmacovigilance: A review. Journal of biomedical informatics. 2015 Apr 30; 54:202-12.
7. S. Fox and M. Duggan. Health online 2013. Health, 2013.
8. Yom-Tov E, Cox IJ, Lampos V. Learning about health and medicine from Internet data. In Proceedings of the Eighth ACM International Conference on Web Search and Data Mining 2015 Feb 2 (pp. 417-418). ACM.
9. Sarker A, Gonzalez G. Portable automatic text classification for adverse drug reaction detection via multi-corpus training, J Biomed Inform 2015; 53: 196–207.
10. Nikfarjam A, Sarker A, O'Connor K, Ginn R, Gonzalez G. Pharmacovigilance from social media: mining adverse drug reaction mentions using sequence labeling with word embedding cluster features. J Am Med Inform Assoc 2015; 22(2).
11. Sarker A, Gonzalez G. Portable automatic text classification for adverse drug reaction detection via multi-corpus training, J Biomed Inform 2015; 53: 196–207.
12. Ginn R, Pimpalkhute P, Nikfarjam A, Patki A, O'Connor K, Sarker A, et al. Mining Twitter for adverse drug reaction mentions: a corpus and classification benchmark. In: Proceedings of the fourth workshop on building and evaluating resources for health and biomedical text processing; 2014.
13. Patki A, Sarker A, Pimpalkhute P, Nikfarjam A, Ginn R, O'Connor K, et al. Mining adverse drug reaction signals from social media: going beyond extraction. In: Proceedings of BioLinkSig 2014; 2014.
14. Bruffaerts R, Vilagut G, Demyttenaere K, Alonso J, AlHamzawi A, Andrade LH, Benjet C, Bromet E, Bunting B, de Girolamo G, Florescu S. Role of common mental and physical disorders in partial disability around the world. The British Journal of Psychiatry. 2012 Jun 1; 200(6):454-61.
15. Alexopoulos GS, Hoptman MJ, Yuen G, et al. Functional connectivity in apathy of late- life depression: A preliminary study. Journal of Affective Disorders. Jul 2013;149(1- 3):398-405
16. Andreescu C, Tudorascu DL, Butters MA, et al. Resting state functional connectivity and treatment response in late-life depression. Psychiatry Research-Neuroimaging. Dec 2013;214(3):313-321
17. Archives of General Psychiatry, 2005 Jun; 62(6): 617-27
18. Journal of the American Medical Association, 2003; Jun 18; 289(23): 3095-105
19. National Institute of Mental Health, 1998
20. Allgulander, C, Sheehan, DV (2002). Generalized anxiety disorder: Raising the expectations of treatment. Psychopharmacol bull 36 (Suppl. 2): 68-78
21. Davis Perahia et al A randomized double blind comparison of duloxetine and venlafaxine in the treatment of patients with major depressive disorder.
22. Miller M, Matthew Miller, V Pate, S A Swanson, D Azrael. CNS drugs: Antidepressant Class, Age, and the Risk of Deliberate Self-Harm: A Propensity Score Matched Cohort Study of SSRI and SNRI Users in the USA. 01/01/2014; 28(1):79.
23. Ginsberg LD, Oubre AY, Daoud YA. Innovations in clinical neuroscience: L-methylfolate Plus SSRI or SNRI from Treatment Initiation Compared to SSRI or SNRI Monotherapy in a Major Depressive Episode. 01/01/2011; 8(1):19.
24. Evidence-based guidelines for treating depressive disorders with antidepressants: A revision of the 2008 British Association for Psychopharmacology guidelines
25. D. Smith, C. Dempster, J. Glanville, N. Freemantle, I. Anderson. Efficacy and tolerability of venlafaxine compared with selective serotonin reuptake inhibitors and other antidepressants: a meta-analysis. British Journal of Psychiatry, 180 (2002), pp. 396–404

26. S.M. Stahl, R. Entsuah, R.L. Rudolph. Comparative efficacy between venlafaxine and SSRIs: a pooled analysis of patients with depression Biological Psychiatry, 52 (2002), pp. 1166–1174
27. Selective Serotonin Reuptake Inhibitors, Venlafaxine and Duloxetine are associated With in Hospital Morbidity but Not Bleeding or Late Mortality after Coronary Artery Bypass Graft Surgery.
28. Allgulander, C, Hartford, J, Russell, JM, Raskin, JR, Erickson, J, Ball, SG et al (2007) Pharmacotherapy for generalized anxiety dis- order: Results of duloxetine treatment from a pooled analysis of 3 clinical trials. Curr Med Res Opin 23: 1245–1252.
29. Bymaster FP, Dreshfield-Ahmad LJ, Threlkeld PG, Shaw JL, Thompson L, Nelson DL, Hemrick-Luecke SK, Wong DT: Comparative affinity of duloxetine and venlafaxine for serotonin and norepinephrine transporters in vitro and in vivo, human serotonin receptor subtypes, and other neuronal receptors. *Neuropsychopharmacology* 2001, 25(6):871-880.
30. Wong DT, Bymaster FP: Dual serotonin and noradrenaline uptake inhibitor class of anti-depressants: potential for greater efficacy or just hype? *Prog Drug Res* 2002, 58:169-222.
31. Stahl, S. (2013). Stahl's essential pharmacology, 4th ed. Cambridge University Press, New York. p. 305, 308, 309.
32. Stahl, SM; Grady, Moret, Briley (Sep 2005). "SNRIs: their pharmacology, clinical efficacy, and tolerability in comparison with other classes of antidepressants". *CNS Spectrums* 10 (9): 732–747.
33. Bymaster, FP; Lee, Knadler (2005). "The dual transporter inhibitor duloxetine: a review of its preclinical pharmacology, pharmacokinetic profile, and clinical results in depression". *Curr Pharm Des* 11 (12): 1475–93.
34. Wellington K, Perry CM (2001). "Venlafaxine extended-release: a review of its use in the management of major depression" (PDF). CNS Drugs 15 (8): 643–69.
35. Choi H, Varian H. Predicting the present with google trends. Economic Record. 2012 Jun 1; 88(s1):2-9.
36. Insights into what the world is searching for -- the new Google Trends, Yossi Matias, Insights Search, The official Google Search blog, September 28, 2012.
37. Mohebbi M, Vanderkam D, Kodysh J, Schonberger R, Choi H, Kumar S. Google correlate whitepaper. Web document: correlate. googlelabs. com/whitepaper. pdf. Last accessed date: August. 2011 Jun; 1:2011.
38. Vanderkam D, Schonberger R, Rowley H, Kumar S. Nearest Neighbor Search in Google Correlate.
39. Abeed Sarker and Graciela Gonzalez. Portable automatic text classification for adverse drug reaction detection via multi-corpus training. Journal of Biomedical Informatics. 53 (2015) 196-207.
40. We used the stemmer provided by the NLTK toolkit: http://www.nltk.org/
41. We used the Regexp tokenizer provided by the NLTK toolkit: http://www.nltk.org/
42. We used the sentimental analysis tool provided by the http://nlp.stanford.edu/sentiment/
43. We used the collocations finder tool provided by the NLTK toolkit: http://www.nltk.org/